\documentclass[10pt,conference]{IEEEtran}
\IEEEoverridecommandlockouts

\usepackage{cite}
\usepackage{amsmath,amssymb,amsfonts}
\usepackage{graphicx}
\usepackage{xcolor}
\usepackage{textcomp}
\def\BibTeX{{\rm B\kern-.05em{\sc i\kern-.025em b}\kern-.08em
    T\kern-.1667em\lower.7ex\hbox{E}\kern-.125emX}}

\usepackage{booktabs}
\usepackage{makecell}
\usepackage[hidelinks]{hyperref}

\begin{document}

\title{Comparative Study of Neural Surrogate Architectures for Autoregressive Prediction of Internal Battery States}

\author{
\IEEEauthorblockN{
Gihyun Lee\IEEEauthorrefmark{1}\IEEEauthorrefmark{2}, 
Thorben Menne\IEEEauthorrefmark{2}, 
Simon Olma\IEEEauthorrefmark{2},
Jakob Hilgert\IEEEauthorrefmark{2},
and Sangyoung Park\IEEEauthorrefmark{1}
}
\IEEEauthorblockA{\IEEEauthorrefmark{1}Technische Universität Berlin, Berlin, Germany\\
Email: \{lee.4, sangyoung.park\}@campus.tu-berlin.de}
\IEEEauthorblockA{\IEEEauthorrefmark{2}IAV GmbH, Berlin, Germany\\
Email: \{thorben.menne, simon.olma, jakob.hilgert\}@iav.de}
}

\maketitle

\begin{abstract}
The Doyle-Fuller-Newman (DFN) model resolves internal electrochemical states in lithium-ion batteries with high fidelity.
However, the numerical solution of its governing equations is computationally prohibitive for real-time deployment, limiting scalability from individual cells to pack and fleet-scale applications.
While machine learning surrogates can substantially reduce inference latency through GPU acceleration, most existing approaches learn solution approximations tied to specific operating conditions rather than learning generalizable state-evolution dynamics.
This work presents a systematic comparison of four neural network architectures (MLP, ResNet, U-Net, FNO) formulated as autoregressive state-transition operators that predict full DFN internal states across a wide range of operating conditions.
To ensure a controlled architectural comparison, all models are trained under a unified framework using multi-step unrolling and current-conditioning, isolating the impact of spatial inductive bias.
Results demonstrate that the U-Net's multi-scale feature hierarchy achieves a mean final-step nRMSE of 3\% averaged across all internal state variables after 300-step autoregressive rollouts, while providing a 5.38$\times$ speed-up over the numerical solver.
These findings highlight spatial inductive bias as a critical determinant of surrogate performance, advancing the development of surrogates for internal state observability for next-generation battery management systems and digital twins.
\end{abstract}

\begin{IEEEkeywords}
lithium-ion batteries, battery management systems, digital twin, electrochemical modeling, Doyle-Fuller-Newman model
\end{IEEEkeywords}

\section{Introduction} \label{sec:introduction}
Lithium-ion batteries (LIBs) are the dominant energy storage technology for electric vehicles (EVs) and grid-scale systems.
Maximizing their performance and lifespan while ensuring safe operation requires accurate estimation of internal electrochemical states \cite{li2021model}.
The Doyle-Fuller-Newman (DFN) model \cite{doyle1993modeling} provides this capability by solving equations governing electrochemical transport and reaction kinetics, yielding high-fidelity internal states. 
However, the computational cost of numerically solving these governing equations is prohibitive for real-time deployment, a challenge that is further compounded when scaling from individual cells to the hundreds or thousands of cells managed by battery management systems (BMS) and monitored by fleet-scale digital twins \cite{hussain2025comprehensive}.

Machine learning (ML) surrogates offer a computationally efficient alternative, exploiting GPU acceleration to substantially reduce per-cell inference latency.
However, most data-driven battery models predict system-level quantities such as voltage, rather than complete spatiotemporal internal states \cite{ji2024review}.
Even among surrogates that target internal states, most approximate specific solution trajectories rather than learning generalizable state-transition dynamics, limiting their applicability to operating conditions seen during training and preventing prediction from arbitrary intermediate states.

Motivated by these limitations, this work presents a systematic comparison of four neural network architectures (MLP, ResNet, U-Net, FNO) formulated as autoregressive state-transition operators for emulating the full-state dynamics of the DFN model.
To ensure a controlled comparison, all architectures are evaluated under a unified training framework, isolating the impact of spatial inductive bias on predictive fidelity, long-horizon stability, and computational efficiency.
These findings advance the development of high-fidelity ML surrogates for internal state observability in next-generation BMS and digital twins.

\section{Theoretical Background} \label{sec:theory}
\begin{figure}[htbp]
    \centering
    \includegraphics[width=0.949\linewidth]{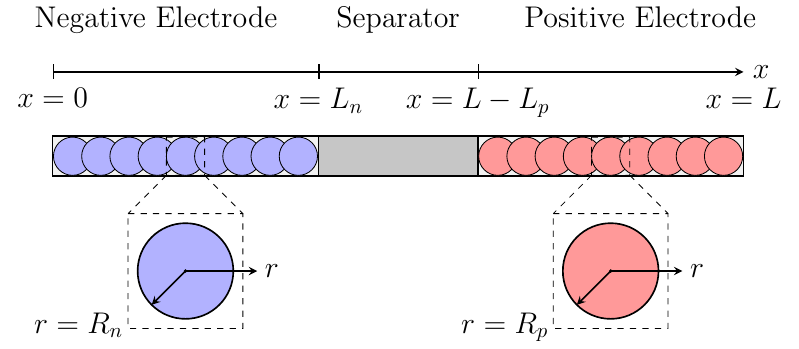}
    \caption{Schematic of the DFN model geometry. Top: Macroscopic domain along the cell thickness (\(x\)) spanning the electrodes and separator. Bottom: Microscopic domain along the radial coordinate (\(r\)) within solid particles.}
    \label{fig:dfn_geometry}
\end{figure}

The DFN model, also known as the Pseudo-Two-Dimensional (P2D) model, is grounded in porous electrode theory and serves as the ground truth for this study \cite{ali2024comparison}.
As illustrated in Fig.~\ref{fig:dfn_geometry}, it couples macroscopic transport along the cell thickness $x$ with microscopic diffusion along the radial coordinate $r$ within spherical solid particles.

The governing dynamics of the DFN model form a system of partial differential-algebraic equations (PDAEs). The differential component comprises conservation laws for mass and charge in the solid and electrolyte phases. Lithium transport within solid particles follows Fick's second law in spherical coordinates \eqref{eq:solid_mass}, while lithium-ion transport in the electrolyte is described by concentrated solution theory \eqref{eq:electrolyte_mass}.
Charge conservation is enforced via Ohm's law in the solid phase \eqref{eq:solid_charge} and via ionic migration and diffusion in the electrolyte \eqref{eq:electrolyte_charge} \cite{ali2024comparison,marquis2019asymptotic}:
\begin{subequations}
\label{eq:dfn_system}
\begin{align}
    \frac{\partial c_{s,k}}{\partial t} &= \frac{1}{r^2} \frac{\partial}{\partial r} \left( D_{s,k} r^2 \frac{\partial c_{s,k}}{\partial r} \right), \label{eq:solid_mass} \\
    \varepsilon_k\frac{\partial c_{e,k}}{\partial t} &= \frac{\partial}{\partial x} \left( D_{e,k}^{\text{eff}} \frac{\partial c_{e,k}}{\partial x} \right) + \frac{1-t^+}{F} a_k j_k, \label{eq:electrolyte_mass} \\
    \frac{\partial}{\partial x} \left( \sigma_k^{\text{eff}} \frac{\partial \phi_{s,k}}{\partial x} \right) &= a_k j_k, \label{eq:solid_charge} \\
    \frac{\partial}{\partial x} \left( \kappa_{e,k}^{\text{eff}} \frac{\partial \phi_{e,k}}{\partial x} \right) &= \frac{\partial}{\partial x} \left( \kappa_{e,k}^{\text{eff}} \frac{2 R T (1-t^+)}{F} \frac{\partial \ln c_{e,k}}{\partial x} \right) \nonumber \\
    & \qquad {} - a_k j_k. \label{eq:electrolyte_charge}
\end{align}
\end{subequations}
Here, subscripts $s$ and $e$ denote the solid and electrolyte phases, respectively, while $k \in \{n,s,p\}$ indexes the spatial regions (negative electrode, separator, positive electrode).
The state variables $c_{s,k}$, $c_{e,k}$, $\phi_{s,k}$, and $\phi_{e,k}$ represent the lithium concentrations and electric potentials in each phase, with solid-phase variables defined only in the electrode regions $k \in \{n,p\}$.
The electrode microstructure is characterized by the porosity $\varepsilon_k$ and specific interfacial area $a_k$.
Transport is governed by the solid diffusivity $D_{s,k}$, cation transference number $t^+$, and effective transport parameters $D_{e,k}^{\text{eff}}$, $\sigma_k^{\text{eff}}$, $\kappa_{e,k}^{\text{eff}}$.

The algebraic component couples the solid and electrolyte phases through the interfacial current density $j_k$, governed by the Butler-Volmer equation \eqref{eq:bv_density}. This reaction rate depends on the local overpotential $\eta_k$ \eqref{eq:bv_eta}, defined as the deviation from the equilibrium open-circuit potential $U_k$, and the exchange current density $i_{0,k}$ \eqref{eq:bv_exchange} \cite{marquis2019asymptotic}:
\begin{subequations}
\label{eq:butler_volmer_system}
\begin{align}
    j_k &= i_{0,k} \sinh\left( \frac{F \eta_k}{2 R T} \right), \label{eq:bv_density} \\
    \eta_k &= \phi_{s,k} - \phi_{e,k} - U_{k}(c_{s,k}^{\mathrm{surf}}), \label{eq:bv_eta} \\
    i_{0,k} &= k_k \sqrt{c_{e,k} \, c_{s,k}^{\mathrm{surf}} \, (c_{s,k}^{\text{max}} - c_{s,k}^{\mathrm{surf}})}, \label{eq:bv_exchange}
\end{align}
\end{subequations}
where $k_k$ is the reaction rate constant, $c_{s,k}^{\text{max}}$ the maximum solid-phase concentration, and $c_{s,k}^{\text{surf}}$ the solid-phase concentration at the particle surface ($r=R_k$).

Together, these transport and reaction mechanisms yield a PDAE system spanning heterogeneous spatial scales, whose stiff, nonlinear dynamics necessitate computationally intensive numerical solvers, motivating the development of the surrogate modeling approach presented in this work.

\section{ML Surrogates for the DFN Model}\label{sec:problem}
\subsection{Problem Formulation}
The DFN model's spatiotemporal dynamics can be reformulated as a discrete-time state-transition system.
At each timestep $t$, the electrochemical state is represented by the state vector $\mathbf{s}_t$:
\begin{align}
    \mathbf{s}_t = \big[ c_e(x), c_{s,k}(x,r), \phi_e(x), \phi_{s,k}(x)\big]^\top, \quad k \in \{n, p\}.
    \label{eq:state_vector}
\end{align}

Since $\mathbf{s}_t$ fully characterizes the electrochemical system at time $t$, its future evolution is determined by the current state and the applied current $I_t$. The DFN dynamics are thus expressed as a nonlinear state-transition function $f$:
\begin{equation}
    \mathbf{s}_{t+1} = f(\mathbf{s}_t, I_t).
    \label{eq:dfn_dynamics}
\end{equation}

The learning task is to approximate this mapping using a parametric function $f_{\theta}$ with learnable parameters $\theta$ such that:
\begin{equation}
    \hat{\mathbf{s}}_{t+1} = f_{\theta}(\hat{\mathbf{s}}_t, I_t).
    \label{eq:surrogate_dynamics}
\end{equation}

This formulation provides three key capabilities for practical deployment in real-time state estimation and monitoring.
First, the model predicts the complete internal state vector $\mathbf{s}_t$, providing full observability of the spatially resolved electrochemical system.
Second, it generalizes across varying initial conditions and current profiles by learning the state-transition dynamics from a diverse range of operating conditions.
Third, it enables solver-free autoregressive operation, supporting efficient long-horizon rollouts without invoking the numerical DFN solver at each timestep.

\subsection{Related Work}
ML surrogates for the DFN model have been proposed across three broad architectural paradigms: CNNs, RNNs, and PINNs, each with distinct limitations that motivate the present work.

McKay et al. \cite{mckay2026learning} proposed a CNN-based surrogate that approximates DFN dynamics by mapping solid-phase concentration fields and terminal voltage to their subsequent state.
While this architecture provides spatially resolved electrode concentrations $c_{s,k}$, it omits the electrolyte concentration $c_e$ and spatially resolved potentials $\phi_e, \phi_{s,k}$.
Since local potentials are the fundamental drivers of Butler-Volmer kinetics, this omission prevents accurate characterization of electrochemical reaction rates.

Recurrent architectures have been proposed to emulate DFN dynamics.
Li et al. \cite{li2021physics} and Mirzaee et al. \cite{mirzaee2023estimation} used specialized LSTM networks to map external signals (current, voltage, and temperature) to internal states.
However, these architectures approximate entire solution trajectories rather than learning the state-transition function.
Consequently, they require the full temporal history from the initial state, preventing prediction from arbitrary intermediate states and limiting their applicability to operating scenarios covered by the training trajectories.
Huang et al. \cite{huang2024minn} addressed this limitation by embedding the governing DAEs into a Model-Integrated Neural Network (MINN).
Nevertheless, this approach requires an embedded numerical solver for time integration, inheriting the computational stiffness of the DFN solver and failing to achieve solver-free autoregressive operation.

PINNs approximate solution fields by incorporating governing equation residuals into the loss function.
When applied to the DFN model, the multi-scale and stiff dynamics result in ill-conditioned optimization, which Hassanaly et al. \cite{hassanaly2024pinn} and Lee et al. \cite{lee2025forward} addressed through stabilization techniques. 
Despite these training advances, PINN formulations approximate spatiotemporal solution fields for fixed operating conditions rather than learning state-transition operators.
This prevents autoregressive prediction from arbitrary intermediate states and requires retraining to adapt to novel current profiles.

This review reveals that no existing approach simultaneously provides complete internal state observability, generalization across operating conditions, and solver-free autoregressive operation. 
Moreover, the relative effectiveness of different architectural inductive biases for capturing the coupled, multi-scale electrochemical dynamics has not been systematically investigated.
This work addresses both limitations by evaluating four architectures commonly employed for PDE surrogate modeling, spanning distinct spatial processing strategies: fully-connected (MLP), local convolutional (ResNet), hierarchical multi-scale (U-Net), and global spectral (FNO).

\section{Methodology}\label{sec:methodology}
\subsection{Unified State Representation}\label{sec:state_representation}
The DFN state vector $\mathbf{s}_t$ comprises six variables with heterogeneous spatial discretizations. Electrolyte fields ($c_e$, $\phi_e$) span the entire cell and are discretized at 60 nodes along the cell thickness.
Solid-phase potentials ($\phi_{s,n}$, $\phi_{s,p}$) are discretized at 20 nodes along the thickness of their respective electrodes.
Solid-phase concentrations ($c_{s,n}, c_{s,p}$) are resolved as 2D fields at $20 \times 20$ nodes per electrode, spanning both electrode thickness and particle radius.

To enable architecture-agnostic processing, each variable is transformed into a $20 \times 20$ spatial grid, yielding a unified six-channel representation $\mathbf{s}_t \in \mathbb{R}^{6 \times 20 \times 20}$. 
Solid-phase concentrations retain their native $20 \times 20$ structure, while solid-phase potentials are replicated along the radial dimension, reflecting their radial uniformity at each electrode position. 
Electrolyte fields are interpolated from 60-node 1D profiles to $20 \times 20$ grids using bilinear interpolation. 

To account for the disparate physical scales of the state variables, per-variable z-score normalization is applied. 
For each variable, normalization statistics (mean and standard deviation) are computed independently at each spatial position across all training samples, preserving spatial gradients while standardizing local magnitudes across variables.

\subsection{Surrogate Architectures}
Each architecture implements the state-transition mapping $\mathbf{s}_t \to \mathbf{s}_{t+1}$ using a distinct spatial processing strategy.
To enable a fair comparison, all models are configured with comparable parameter counts ($\sim$3M).
Table~\ref{tab:architecture_hyperparameters} summarizes their hyperparameter configurations, and each architecture's spatial processing mechanism is detailed below.

\begin{table}[htbp]
    \caption{Hyperparameter Configurations for Surrogate Architectures}
    \centering
    \footnotesize
    \renewcommand{\arraystretch}{1.2}
    \begin{tabular}{lccc}
        \toprule
        \textbf{Architecture} & \textbf{Hidden Dims} & \textbf{Depth} & \textbf{Key Params} \\
        \midrule
        MLP    & 768$\to$1152$\to$768 & 3 Hidden Layers & --- \\
        ResNet & 48$\to$96$\to$192    & 3 Residual Blocks & $3{\times}3$ Kernel \\
        U-Net  & 80$\to$160$\to$320   & 2 Stages & $3{\times}3$ Kernel \\
        FNO    & 64                   & 4 Fourier Layers & 10 Modes \\
        \bottomrule
        \multicolumn{4}{l}{\footnotesize All models use GELU activation and batch normalization.} \\
    \end{tabular}
    \vspace{-18pt}
    \label{tab:architecture_hyperparameters}
\end{table}

\subsubsection{MLP}
The MLP employs a spatially agnostic architecture.
Rather than operating on the unified $20\times20$ grid, the MLP processes the native DFN discretization directly. All state variables are flattened and concatenated into a single input vector and processed through fully-connected layers. 
While dense connectivity enables global information exchange, the MLP lacks explicit spatial structure, requiring it to learn the spatial dependencies purely from data.

\subsubsection{ResNet}
ResNet \cite{he2016deep} processes the unified state representation through sequential residual blocks, where identity shortcuts facilitate incremental feature refinements.
This architecture maintains uniform spatial resolution throughout, applying localized $3\times3$ convolutions at every stage.
The cumulative receptive field of the network spans the full $20\times20$ grid, with spatial context assembled incrementally through network depth.

\subsubsection{U-Net}
The U-Net \cite{ronneberger2015u} employs a hierarchical encoder-decoder architecture with multi-scale processing. 
Progressive downsampling compresses the spatial representation into a bottleneck that captures global context across the full $20\times20$ grid.
Lateral skip connections then reintroduce high-resolution features into the decoder, preserving sharp spatial gradients. 
This design enables the network to capture long-range dependencies and local spatial variations.

\subsubsection{FNO}
The FNO \cite{li2021fourier} operates in the frequency domain by applying Fourier transforms to the spatial representation before performing spectral convolutions.
This approach provides a global receptive field within each layer, as operations in frequency space affect all spatial positions simultaneously.
The architecture retains a limited number of Fourier modes, capturing predominantly low-frequency spatial structure.

\subsection{Current Conditioning Methods}
As the applied current $I_t$ drives electrochemical kinetics and transport, the surrogate model must be explicitly conditioned on this control input.
Three distinct conditioning paradigms are considered for integrating the control input: input concatenation, Feature-wise Linear Modulation (FiLM), and Conditional Instance Normalization (CIN). 
While MLP uses input concatenation exclusively, spatial architectures (ResNet, U-Net, and FNO) are evaluated across all three methods, with architecture-specific optimal configurations identified empirically in Section~\ref{sec:current_conditioning} and applied in all subsequent experiments.

\subsubsection{Input Concatenation}
The scalar current $I_t$ is concatenated with the input representation, either directly to the flattened vector (MLP) or broadcast as an additional spatial channel (ResNet, U-Net, FNO).
This provides the model with direct access to the control input but offers no explicit mechanism for modulating intermediate feature representations.

\subsubsection{FiLM}
FiLM \cite{perez2018film} conditions the network by applying learned affine transformations to intermediate feature maps $\mathbf{h} \in \mathbb{R}^{C \times H \times W}$. 
Based on the applied current $I_t$, the mechanism computes channel-wise scale $\gamma(I_t)$ and shift $\beta(I_t)$ parameters:
\begin{equation}
    \text{FiLM}(\mathbf{h}, I_t) = \gamma(I_t) \odot \mathbf{h} + \beta(I_t),
\end{equation}
where $\gamma, \beta: \mathbb{R} \to \mathbb{R}^C$ are learned linear projections.
Applied after convolution blocks, this modulation enables dynamic adaptation of internal representations in response to the control input.

\subsubsection{CIN}
CIN \cite{dumoulin2017learned} is adapted for continuous conditioning by replacing discrete style embeddings with learned functions of $I_t$.
This approach standardizes feature maps $\mathbf{h}$ via instance normalization before applying affine modulation:
\begin{equation}
    \text{CIN}(\mathbf{h}, I_t) = \gamma(I_t) \odot \frac{\mathbf{h} - \mu(\mathbf{h})}{\sigma(\mathbf{h})} + \beta(I_t),
\end{equation}
where $\mu(\mathbf{h}), \sigma(\mathbf{h})\in \mathbb{R}^C$ are channel-wise instance statistics, and $\gamma, \beta$ are linear projections identical to FiLM. 
By decoupling internal feature scales from the conditioning signal, CIN provides a normalized representation before applying the learned modulation.

\subsection{Temporal Training Strategy}
During autoregressive inference, the surrogate model recursively uses its own predictions as inputs, causing prediction errors to accumulate over time. 
Training on single-step predictions with ground-truth states creates a distributional mismatch, known as exposure bias, as the model never encounters its own errors during training, which can lead to instability in long-horizon rollouts\cite{brandstettermessage}.

To address this, unrolled training over $K$ consecutive timesteps is employed.
Starting from a ground-truth state $\mathbf{s}_t$, the model autoregressively predicts the next $K$ states, and the multi-step prediction loss is computed as:
\begin{equation}
    \mathcal{L}_{\text{multi-step}} = \frac{1}{K} \sum_{k=1}^{K} \|\mathbf{s}_{t+k} - \hat{\mathbf{s}}_{t+k}\|_2^2,
    \label{eq:multi_loss}
\end{equation}
where $\hat{\mathbf{s}}_{t+k}$ is obtained by recursively applying Eq.~\eqref{eq:surrogate_dynamics}.
By backpropagating through the unrolled sequence, the model is trained to handle its own prediction errors as inputs, improving robustness and long-horizon stability.
As detailed in Section~\ref{sec:unroll_horizon}, an unroll horizon of $K=10$ balances long-term prediction stability with computational cost.

\section{Experimental Setup}\label{sec:experimentalsetup}
\subsection{Dataset}
\subsubsection{Simulation Setup}
A dataset for training and evaluation is generated using the isothermal DFN model implemented within the PyBaMM framework \cite{sulzer2021python}. 
The model uses the Marquis2019 parameter set \cite{marquis2019asymptotic}, representing a LIB cell with a graphite (MCMB 2528) negative electrode and LiCoO$_2$ positive electrode.
The spatial domain is discretized using 20 finite volumes per macroscopic region (negative electrode, separator, positive electrode) and 20 radial nodes within each electrode particle, consistent with the state representation in Section~\ref{sec:state_representation}.
The resulting DAE system is solved using the CasADi solver and sampled at 1\,Hz.
\begin{table}[htbp]
    \caption{Summary of Current Excitation Profiles and Operating Conditions for DFN Dataset Generation}    
    \centering
    \renewcommand{\arraystretch}{1.05}
    \resizebox{\columnwidth}{!}{
    \footnotesize
        \begin{tabular}{lcccc}
            \toprule
            \textbf{Profile Type} & \textbf{C-Rate Range} & \textbf{Initial SOC} & \textbf{Duration} & \textbf{Count} \\
            \midrule
            \multicolumn{5}{l}{\textit{Synthetic Profiles}} \\
            \quad CC Charge & [$-$3.0C, $-$1.0C] & [0\%, 80\%] & 900\,s & 16 \\
            \quad CC Discharge & [1.0C, 3.0C] & [20\%, 100\%] & 900\,s & 16 \\
            \quad Sinusoidal & $\pm$2.0C, $\pm$3.0C & [10\%, 100\%] & 600\,s & 40 \\
            \quad Pulse-Relaxation & $\pm$1.5C, $\pm$2.5C & [10\%, 100\%] & 600\,s & 20 \\
            \midrule
            \multicolumn{5}{l}{\textit{Driving Cycles}} \\
            \quad WLTC & [$-$0.69C, 1.45C] & [10\%, 100\%] & 1800\,s & 10\\
            \quad UDDS & [$-$0.56C, 1.01C] & [10\%, 100\%] & 1368\,s & 10 \\
            \quad US06 & [$-$1.68C, 3.24C] & [10\%, 100\%] & 600\,s & 10 \\
            \midrule
            \textbf{Total} & \textbf{[$-$3.00C, 3.24C]} & \textbf{[0\%, 100\%]} & \textbf{28\,h} & \textbf{122} \\
            \bottomrule
        \end{tabular}
        }
    \label{tab:dataset_generation}
    \vspace{-10pt}
\end{table}

\subsubsection{Data Generation}
To facilitate generalization across a wide operating range, a dataset of 122 trajectories spanning C-rates from $-$3.0C to 3.24C and initial SOC values from 0\% to 100\% is generated (Table~\ref{tab:dataset_generation}).
The dataset combines standardized driving cycles (WLTC, UDDS, US06) representing realistic EV operation with synthetic profiles designed to systematically probe distinct electrochemical regimes.
Specifically, constant current (CC) profiles capture sustained load dynamics, sinusoidal profiles with periods of 50\,s and 100\,s excite dynamic transient responses, and pulse-relaxation sequences (10\,s pulse, 40\,s rest) characterize relaxation behavior. 
Each profile type is simulated across multiple initial SOC values and C-rates, yielding the DFN state vector $\mathbf{s}_t$ at 1\,Hz and ensuring comprehensive coverage of the battery's operating envelope.

\subsubsection{Data Partitioning}
To prevent temporal data leakage while maintaining distributional balance, a stratified trajectory-based split is performed.
All timesteps from each simulation trajectory are assigned to the same partition, ensuring no temporal overlap between sets.
Stratification across 5 SOC $\times$ 20 C-rate bins ensures balanced coverage of the operating space.
This yields training (83 trajectories, 69k samples), validation (19 trajectories, 15k samples), and test (20 trajectories, 16k samples) sets.

\subsection{Training Details}
All models are implemented in PyTorch and trained to minimize the multi-step prediction loss (Eq.~\eqref{eq:multi_loss}) with $K{=}10$ rollout steps.
Optimization is performed using the AdamW optimizer with weight decay $\lambda = 1\times10^{-4}$ and a batch size of 128. 
The learning rate follows a cosine annealing schedule, decaying from $\eta_0 =8\times 10^{-4}$ to $\eta_{\text{min}} = 10^{-6}$ over 200 epochs. 
Gradient norm clipping with threshold 2.0 ensures numerical stability during backpropagation through the unrolled sequence. 
Training inputs are augmented with Gaussian noise  ($\sigma = 1.5\%$ of each variable's range) to improve robustness to autoregressive error accumulation.
Training is terminated using early stopping based on validation loss with patience of 40 epochs.
All experiments are conducted on a NVIDIA A100 40\,GB GPU (3g.20\,GB MIG instance).

\subsection{Evaluation Metrics}
Prediction accuracy is quantified using normalized root mean squared error (nRMSE). 
At each rollout step $k$, RMSE is computed by first averaging squared errors over spatial grid points for each sample, then averaging across all test samples.
The result is normalized by the global range (max $-$ min) of each ground-truth variable to enable comparison across disparate physical scales.
To evaluate autoregressive stability, 300-step rollouts are initiated from every valid starting point in the test set, yielding nRMSE as a function of rollout step $k$ that quantifies error accumulation.
A global metric is obtained by averaging nRMSE over all rollout steps. 
For system-level evaluation, RMSE for terminal voltage and SOC are reported for CC discharge at 1C, 2C, and 3C. 
These quantities are derived from predicted internal states: terminal voltage from electrode potential differences at the current collectors, and SOC from the volume-averaged negative electrode concentration.

Computational efficiency is assessed through inference latency and model complexity.
The PyBaMM solver, executed sequentially (one timestep per call) on a single AMD EPYC 7742 CPU core, serves as the baseline.
Surrogate inference latency is measured on an NVIDIA A100 40\,GB GPU with batch size 1, representing single-sample online inference.
Reported timings reflect model inference only, excluding data transfer between CPU and GPU, and are averaged over 100 trajectories.
Model complexity is quantified by parameter count and peak GPU memory consumption during inference.

\section{Results and Discussion}\label{sec:results}
\subsection{State Evolution and Autoregressive Rollout Performance}
\begin{figure*}[htbp]
    \centering
    \includegraphics[width=1.0\linewidth]{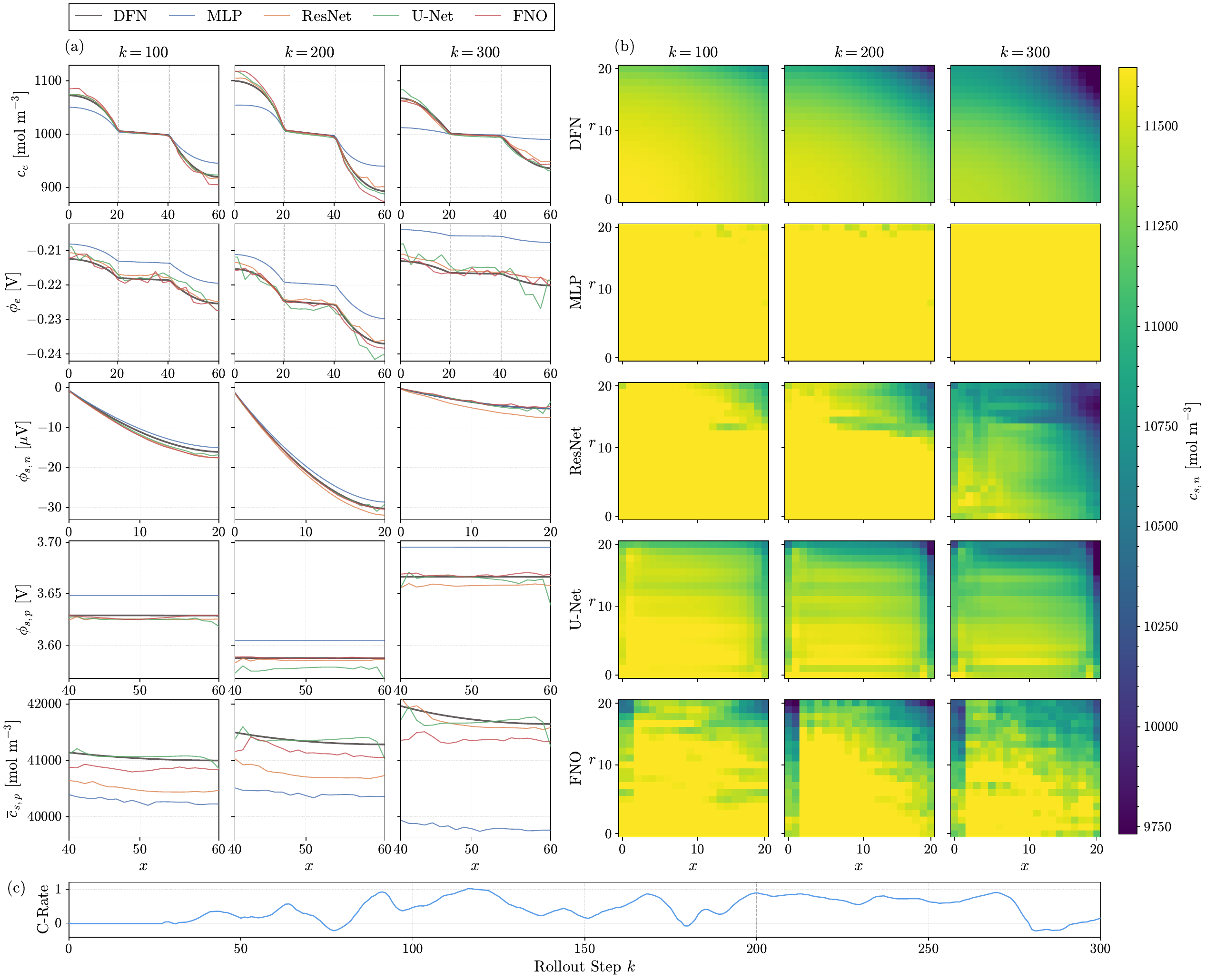}
    \vspace{-20pt}
    \caption{Spatiotemporal evolution of internal states during 300-step autoregressive rollout under a WLTC driving cycle. (a) 1D spatial profiles at steps $k =100, 200, 300$, showing electrolyte concentration ($c_e$) and potential ($\phi_e$) across the full cell, electrode potentials ($\phi_{s,n},\phi_{s,p}$), and radially-averaged positive electrode concentration ($\bar{c}_{s,p}$). (b) 2D negative electrode concentration ($c_{s,n}$) fields at steps $k =100, 200, 300$, resolved over electrode thickness ($x$) and particle radius ($r$). (c) Applied C-rate profile over the 300-step rollout, representative of WLTC driving cycle dynamics.}
    \label{fig:full_system_evolution}
    \vspace{-9pt}
\end{figure*}

Spatiotemporal evolution of internal states over a 300-step WLTC rollout (Fig.~\ref{fig:full_system_evolution}) reveals architecture-dependent performance differences that vary across state variables.
For 1D macroscopic profiles (Fig.~\ref{fig:full_system_evolution} (a)), ResNet, U-Net, and FNO maintain close agreement with the DFN reference, while the MLP exhibits significant error accumulation.
Concentration fields, both radially-averaged positive electrode (Fig.~\ref{fig:full_system_evolution} (a), row 5) and spatially-resolved negative electrode (Fig.~\ref{fig:full_system_evolution} (b)), expose the largest architectural disparities.
The MLP exhibits catastrophic spatial collapse (Fig.~\ref{fig:full_system_evolution} (b), row 2), producing spatially uniform predictions due to its inability to exploit spatial relationships.
The ResNet captures surface concentration depletion at steps $k=100,200$, but fails to propagate gradients into the particle interior due to its incremental spatial context assembly through network depth.
While predictions converge by $k=300$, this transient lag could underestimate diffusion limitations during rapid load changes.
The FNO shows intermediate performance, with spatial artifacts near the current collector (Fig.~\ref{fig:full_system_evolution} (b), row 5, $x\approx0$) and underestimated surface lithium depletion at step $k=300$ ($r=R_k$). 
These limitations stem from the truncation of higher Fourier modes, which acts as a low-pass filter that effectively captures smooth global variations but attenuates the high-frequency content required for sharp spatial gradients. 
The U-Net maintains close agreement with DFN throughout, accurately resolving both surface depletion and radial gradient propagation into the particle core. Despite minor spatial discontinuities relative to the smooth DFN reference, its multi-scale convolutional features effectively capture the coupled diffusion dynamics.

\begin{figure*}[htbp]
    \centering
    \includegraphics[width=0.8\linewidth]{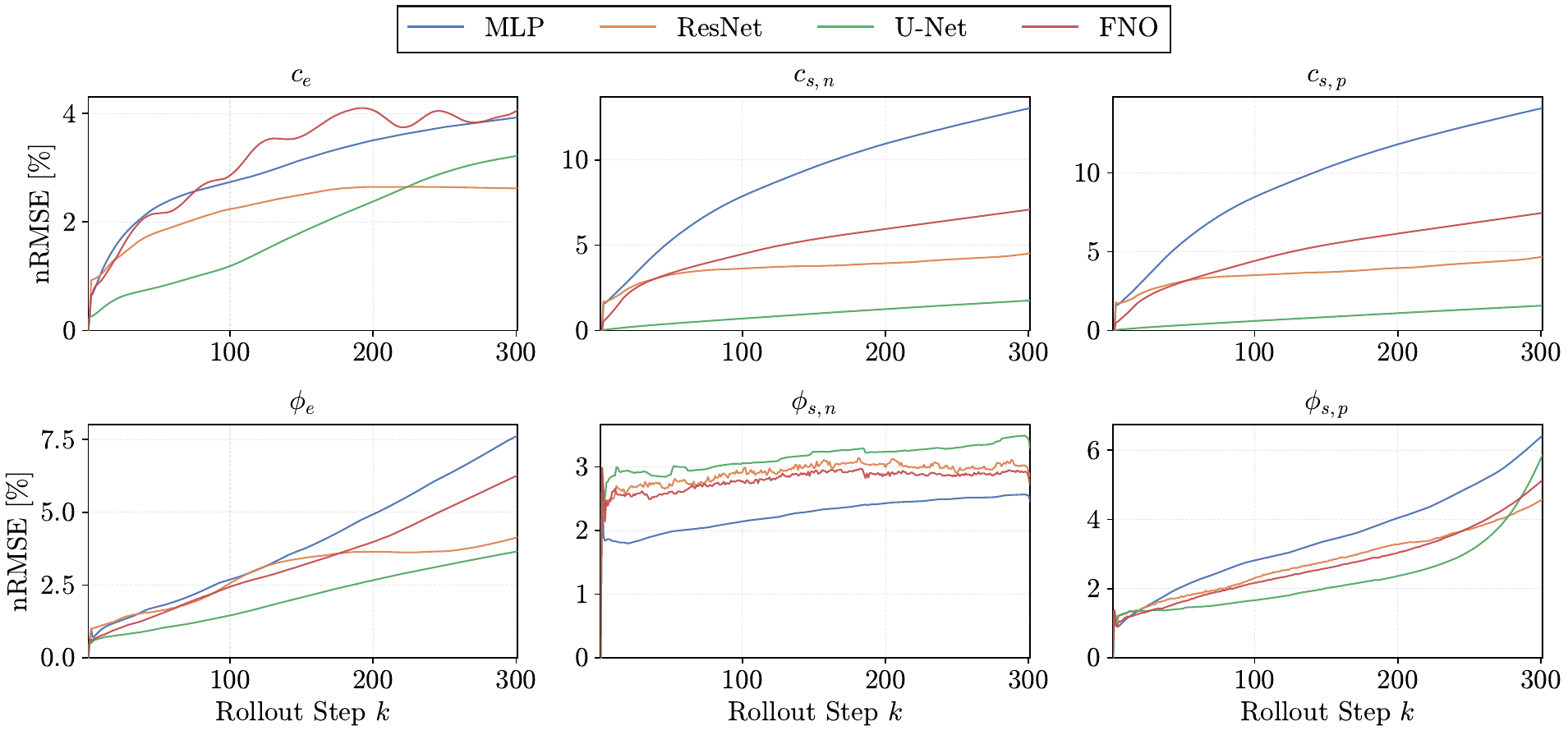}
    \vspace{-5pt}
    \caption{Evolution of nRMSE over 300-step autoregressive rollout for each internal state variable. Errors are spatially averaged over grid points and averaged across all test trajectories. Each panel shows a different internal state variable ($c_e$, $c_{s,n}$, $c_{s,p}$, $\phi_e$, $\phi_{s,n}$, $\phi_{s,p}$).}
    \label{fig:error_evolution}
    \vspace{-13pt}
\end{figure*}

Error accumulation over the 300-step rollout (Fig.~\ref{fig:error_evolution}) quantifies the long-term autoregressive stability of each architecture.
The U-Net demonstrates the best overall performance, maintaining gradual error growth and yielding the lowest final-step nRMSE for electrode concentrations ($c_{s,n},c_{s,p}\approx2\%$) and electrolyte potential ($\phi_e\approx3.5\%$).
ResNet achieves the lowest final error for electrolyte concentration ($c_e\approx3\%$) and positive electrode potential ($\phi_{s,p}\approx4.5\%$), demonstrating long-term stability despite initially trailing the U-Net.
The MLP maintains the lowest error for negative electrode potential ($\phi_{s,n}\approx2.5\%$), despite showing rapid error growth in diffusion-dominated concentration fields ($c_{s,n},c_{s,p}>12\%$).
The FNO remains competitive in potential fields ($\phi_{s,n}\approx3\%$, $\phi_{s,}\approx5\%$) where instantaneous equilibrium dominates, but its performance in concentration fields is limited by spectral mode truncation.
These results confirm that multi-scale hierarchical processing (U-Net) most efficiently captures both smooth global fields and sharp localized gradients, whereas architectures optimized for local (ResNet) or global (FNO) features alone exhibit variable-dependent limitations.

\subsection{System-Level Performance}
\begin{figure*}[htbp]
    \centering
    \includegraphics[width=0.90\linewidth]{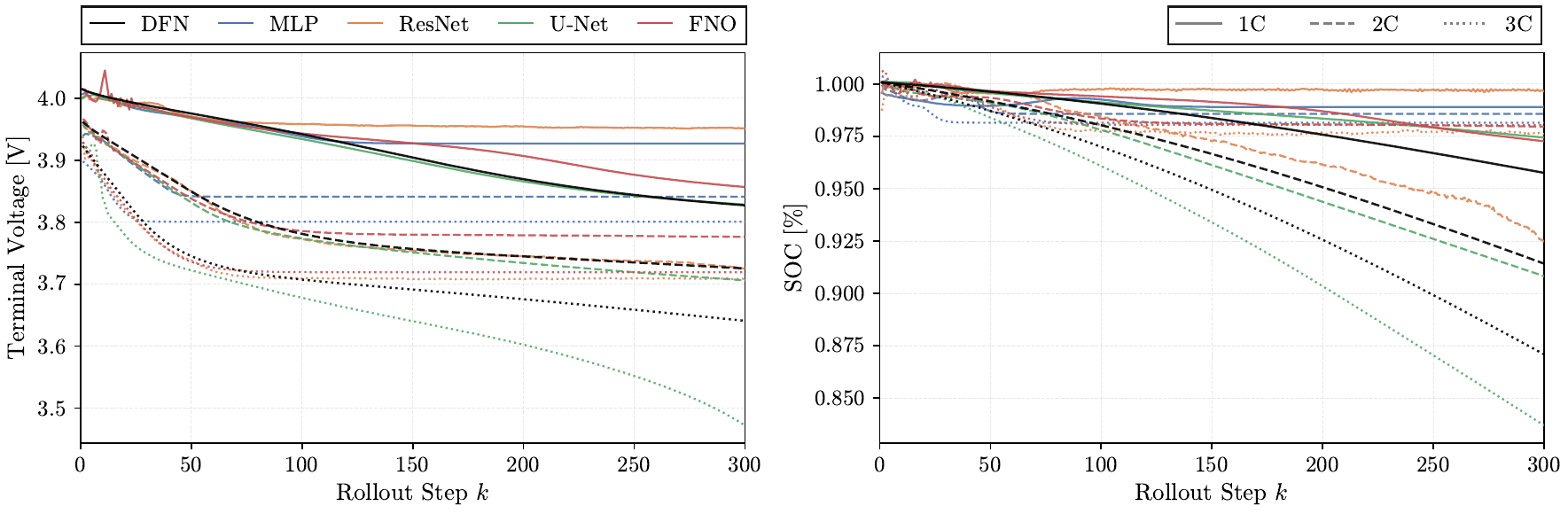}
    \vspace{-5pt}
    \caption{Terminal voltage and SOC predictions during 300-step rollouts at 1C, 2C, 3C CC discharge from fully charged conditions.}
    \label{fig:vol_soc}
    \vspace{-13pt}
\end{figure*}

Terminal voltage and SOC are evaluated to assess whether internal state prediction accuracy translates to reliable system-level predictions.
These quantities are tested under CC discharge at 1C, 2C, and 3C from fully charged conditions (Fig.~\ref{fig:vol_soc}).
The U-Net maintains close alignment with the DFN reference at 1C and 2C, with more notable deviations emerging at 3C.
ResNet exhibits pronounced C-rate dependence, achieving excellent accuracy at 2C but degrading substantially at 1C and 3C.
The FNO maintains reasonable voltage tracking at 1C and 2C but shows large deviations at 3C, while SOC predictions remain accurate only at 1C and drift substantially at higher C-rates.
The MLP exhibits systematic overestimation of voltage and severe SOC drift across all C-rates, confirming its inability to capture electrochemical dynamics.

Table~\ref{tab:vol_soc} quantifies system-level performance by averaging errors from 300-step rollouts initiated at every valid timestep across each discharge trajectory.
The U-Net achieves lowest errors for five of six metrics, with ResNet achieving lowest voltage RMSE at 2C (10.7\,mV).
The U-Net's consistently low errors across all C-rates ($\le$16.1\,mV, $\le$2.23\% SOC) demonstrate robust system-level prediction, while MLP's large errors (33.5-45.6\,mV, 7.9-12.4\% SOC) underscore the critical importance of spatial inductive bias.

\begin{table}[htbp]
  \centering
  \caption{Terminal Voltage and SOC RMSE Under CC Discharge}
  \footnotesize
  \renewcommand{\arraystretch}{1.2}
  \begin{tabular}{lcccccc}
    \toprule
    & \multicolumn{3}{c}{\textbf{Voltage} [mV]} & \multicolumn{3}{c}{\textbf{SOC} [\%]} \\
    \cmidrule(lr){2-4} \cmidrule(lr){5-7}
    \textbf{Model} & \textbf{1C} & \textbf{2C} & \textbf{3C} & \textbf{1C} & \textbf{2C} & \textbf{3C} \\
    \midrule
    MLP    & 33.5 & 45.6 & 36.0 & 7.94 & 12.40 & 9.20 \\
    ResNet & 14.1 & \textbf{10.7} & 35.6 & 2.89 & 3.25 & 10.42 \\
    U-Net  & \textbf{6.9} & 11.7 & \textbf{16.1} & \textbf{1.20} & \textbf{0.73} & \textbf{2.23} \\
    FNO    & 16.7 & 14.0 & 38.4 & 3.76 & 3.92 & 8.59 \\
    \bottomrule
  \end{tabular}
  \label{tab:vol_soc}
\end{table}
\begin{table}[htbp]
    \caption{Computational Efficiency of Surrogate Models}
    \footnotesize
    \centering
    \renewcommand{\arraystretch}{1.15}
        \begin{tabular}{lcccc}
            \toprule
            \textbf{Model} & \textbf{\makecell{Inference\\(ms/step)}} & \textbf{\makecell{Speed-up\\(vs. DFN)}} & \textbf{\makecell{Parameters \\(M)}} & \textbf{\makecell{Peak Memory\\(MB)}} \\
            \midrule
            MLP & 0.45 & $24.16\times$ & 3.25 & 12.44 \\
            ResNet & 3.71 & $2.93\times$ & 2.65 & 10.15 \\
            U-Net & 2.02 & $5.38\times$ & 2.95 & 11.30 \\
            FNO  & 1.90 & $5.72\times$ & 3.30 & 25.07 \\
            \bottomrule
        \end{tabular}
    \label{tab:computational_efficiency}
    \vspace{-13pt}
\end{table}
\subsection{Computational Efficiency}
Table~\ref{tab:computational_efficiency} summarizes the computational requirements of each surrogate, benchmarking inference latency against the PyBaMM DFN solver (10.87\,ms/step on single CPU core), with all surrogates achieving considerable speed-up (2.93--24.16$\times$).
The U-Net achieves the most favorable balance of predictive fidelity with computational efficiency (5.38$\times$ speed-up, 11.30\,MB memory), making it most suitable for real-time deployment.
While the MLP offers the fastest inference (24.16$\times$), its poor accuracy precludes practical deployment.
ResNet achieves the lowest speed-up (2.93$\times$) despite competitive accuracy.
The FNO's high memory footprint (25.07\,MB) may constrain embedded deployment despite competitive speed-up (5.72$\times$). 
All architectures have comparable parameter counts (2.65--3.30\,M), confirming performance differences stem from inductive biases rather than model capacity.
\subsection{Sensitivity Analysis}
\subsubsection{Impact of Current Conditioning Method} \label{sec:current_conditioning}
To assess the effect of current conditioning on prediction accuracy, ResNet, U-Net, and FNO are each trained with three conditioning methods (concatenation, CIN, FiLM).
Table~\ref{tab:ablation_conditioning} reports global nRMSE averaged over 300-step test rollouts, revealing that the optimal conditioning strategy is architecture-dependent.
ResNet benefits most from CIN (3.03\%), where instance-level feature modulation aligns well with its incremental residual learning.
The U-Net achieves the lowest error with FiLM (1.70\%), where channel-wise modulation allows adaptive feature scaling across its multi-scale encoder-decoder structure.
\begin{table}[htbp]
    \caption{Current Conditioning Method Comparison}
    \footnotesize
    \renewcommand{\arraystretch}{1.15}
    \centering
    \setlength{\tabcolsep}{10pt}
        \begin{tabular}{lccc}
            \toprule
            \textbf{Model} & \textbf{Concatenation} & \textbf{CIN} & \textbf{FiLM} \\
            \midrule
            ResNet  & 5.26\% & \textbf{3.03\%} & 5.17\% \\
            U-Net   & 1.76\% & 2.35\% & \textbf{1.70\%} \\
            FNO     & \textbf{3.72\%} & 6.92\% & 4.11\% \\
            \bottomrule
        \end{tabular}
    \label{tab:ablation_conditioning}
    \vspace{-15pt}
\end{table}
Conversely, the FNO performs best with simple concatenation (3.72\%), suggesting that input-level conditioning better preserves the spectral representations than intermediate feature modulation.
The U-Net with FiLM achieves the lowest overall error, establishing it as the optimal configuration for autoregressive DFN surrogate modeling.
\subsubsection{Impact of Training Unroll Horizon}\label{sec:unroll_horizon}
\begin{figure}[htbp]
    \centering
    \includegraphics[width=0.941\linewidth]{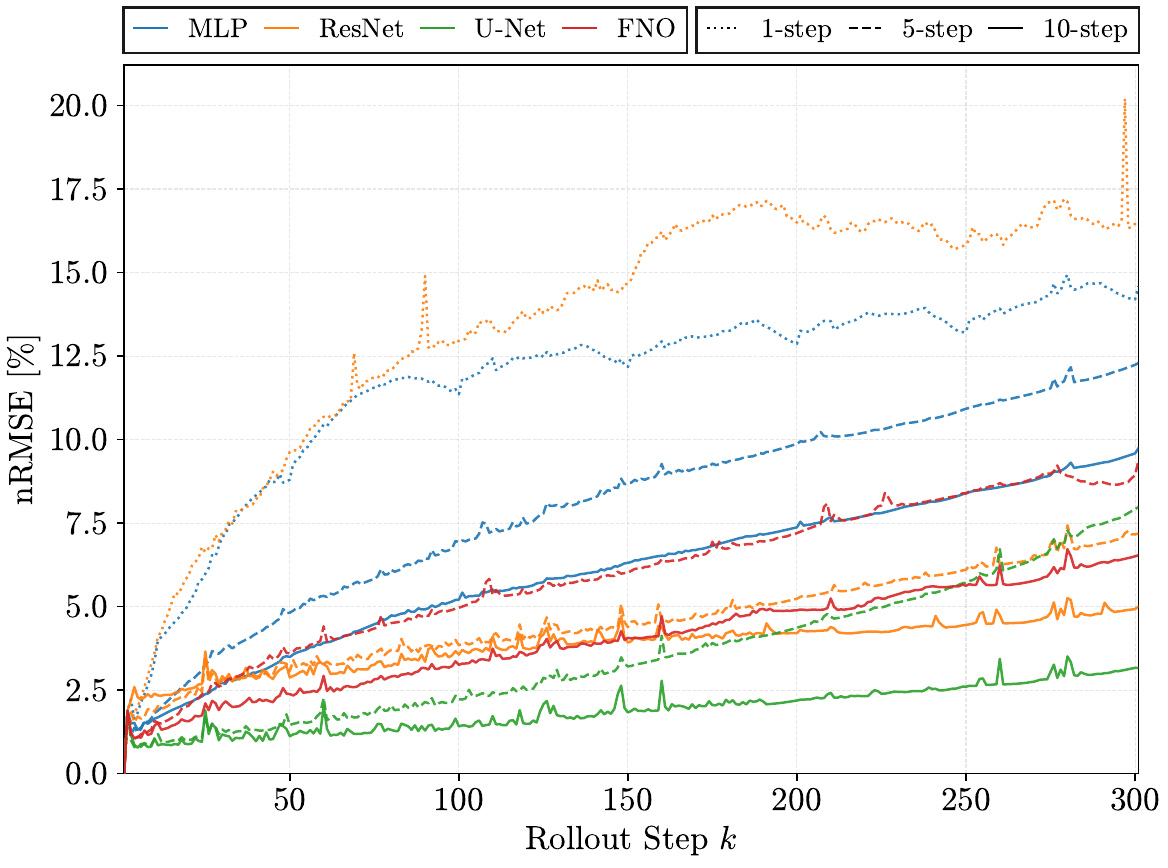}
    \vspace{-5pt}
    \caption{Impact of training unroll horizon $K$ on 300-step rollout stability. Global nRMSE evolution for $K=1, 5, 10$ across all architectures.}
    \label{fig:unroll}
\end{figure}

The impact of training unroll horizon $K$ on rollout stability is analyzed to assess its role in mitigating exposure bias (Fig.~\ref{fig:unroll}).
Increasing $K$ consistently reduces error accumulation across all architectures, with ResNet showing the largest quantifiable improvement (70\% final-step nRMSE reduction from $K=1$ to $K=10$).
Critically, U-Net and FNO trained with $K=1$ diverged and are excluded from the figure, demonstrating that these architectures require multi-step gradients to learn self-stabilizing dynamics.
These results validate the selection of $K=10$ as the optimal operating point, balancing robust long-horizon stability with the computational cost of backpropagation through the unrolled sequence.

\section{Conclusion and Outlook}\label{sec:conclusion}
This work presented a systematic comparison of four neural surrogate architectures (MLP, ResNet, U-Net, FNO) for autoregressive prediction of internal electrochemical states of the DFN model, isolating the effect of spatial inductive bias on predictive fidelity, long-horizon stability, and computational efficiency. The results demonstrate that this spatial processing characteristic critically determines the surrogate performance.

The MLP, lacking explicit spatial structure, exhibits the highest overall error accumulation. ResNet improves stability through local spatial processing but captures coupled dynamics less efficiently due to its gradual information propagation. The FNO captures global interactions through spectral methods but shows degraded performance in resolving localized spatial variations in concentration fields. The U-Net's hierarchical multi-scale architecture bridges these complementary limitations, aligning with the coupled dynamics of the DFN model by representing both globally coupled potentials and locally evolving concentrations. This yields the most favorable trade-off between predictive fidelity and computational efficiency, achieving a mean final-step nRMSE of $3\%$ across all internal state variables after 300-step rollouts, while maintaining a 5.38$\times$ speed-up over the numerical solver.

Beyond architectural choice, two methodological factors are identified as essential for robust autoregressive performance. Multi-step unrolled training significantly reduces error accumulation across all architectures and is necessary to prevent divergence for U-Net and FNO. In addition, the effectiveness of current conditioning is highly architecture-dependent, requiring alignment with each model’s spatial processing characteristics. Since both factors were applied consistently, the observed performance hierarchy is primarily attributable to
differences in spatial inductive bias.

The scope of this study is bounded by the isothermal formulation of the DFN model and the use of a single cell chemistry
and parameterization. While these conditions are sufficient for
a controlled comparison of architectural inductive biases, the
transferability of the observed performance hierarchy across
diverse cell chemistries and extended physical modeling domains remains to be validated. Furthermore, the surrogate
operates in open-loop, requiring accurate initial internal states
that are not directly measurable in practice, while accumulated
errors remain uncorrected throughout the rollout.

Future work will address these limitations by evaluating the
framework across multiple cell chemistries and parameterizations, and extending the methodology to coupled electrothermal dynamics and degradation mechanisms. In parallel, integrating the surrogate into a closed-loop observer framework,
where feedback from measurable system-level quantities such
as terminal voltage iteratively corrects internal state predictions, will reduce sensitivity to initialization uncertainty and
mitigate long-horizon drift. These extensions aim to advance
the practical viability of high-fidelity surrogate models for
real-time internal state observability in next-generation BMS
and digital twins.

\bibliographystyle{IEEEtran}
\bibliography{references}

\end{document}